\documentclass[11pt]{article}

\usepackage[final]{acl}

\usepackage{times}
\usepackage{latexsym}

\usepackage[T1]{fontenc}

\usepackage[utf8]{inputenc}
\usepackage{newunicodechar}
\newunicodechar{−}{-}

\usepackage{microtype}
\usepackage{booktabs}
\usepackage{inconsolata}
\usepackage{multirow}
\usepackage{graphicx}
\usepackage{amssymb}
\usepackage{amsmath}
\usepackage{makecell}
\usepackage{subcaption}
\usepackage{xcolor}
\usepackage{amssymb}

\usepackage[breakable,skins]{tcolorbox}

\newtcolorbox{promptbox}[1][]{
    breakable,
    colback=gray!10,    
    colframe=gray!20,     
    title=#1,           
    fonttitle=\bfseries,
    boxrule=0.5mm,      
    arc=2mm,            
    outer arc=2mm,      
    coltitle=black,     
    enhanced,
}
\title{Modular Expert Merging for Biomedical Retrieval}

\newcommand*\samethanks[1][\value{footnote}]{\footnotemark[#1]}

\author{
\vspace{-1cm}\\
  {\bf Sameh Khattab}\textsuperscript{1,4}\thanks{Corresponding authors: \textit{sameh.khattab@uk-essen.de}, \textit{jcorbeil@microsoft.com}},
  {\bf Jean-Philippe Corbeil}\textsuperscript{2}\samethanks[1],
  {\bf Osman Alperen \c{C}inar-Kora\c{s}}\textsuperscript{1,4}\\
  {\bf Amin Dada}\textsuperscript{1},
  {\bf Julian Friedrich}\textsuperscript{1},
  {\bf Jiawei He}\textsuperscript{3},
  {\bf Douglas Teodoro}\textsuperscript{3},
  {\bf Jens Kleesiek}\textsuperscript{1,4}\thanks{Other affiliations:
 Cancer Research Center Cologne Essen (CCCE), German Cancer Consortium (DKTK, Partner site Essen) and Department of Physics of TU Dortmund, Germany.}\\\\
  \textsuperscript{1}Institute for Artificial Intelligence in Medicine (IKIM), University Hospital Essen, Germany\\\ 
  \textsuperscript{2}Microsoft Healthcare \& Life Sciences, Canada\\
  \textsuperscript{3}Department of Radiology and Medical Informatics, University of Geneva, Switzerland\\
  \textsuperscript{4}Faculty of Computer Science, University of Duisburg-Essen, Germany\ \ 
\vspace{-1.2cm}
}

\begin{document}
\maketitle
\begin{abstract}
Adapting general-purpose LLMs into domain-specialized dense retrievers typically requires large-scale training on mixed-domain data. We show that merging independently trained domain-specialized experts consistently exceeds this approach across four decoder-only LLM families (0.6B–7B), four merging methods, and twelve medical and general retrieval tasks from MTEB, suggesting that parameter-space composition captures complementary domain strengths that large-scale mixed-domain training averages out. To further maximize expert quality, we introduce Synthesize-Train-Merge (STM), a modular framework that synthesizes hard negatives with a top-tier LLM and fine-tunes domain-specialized experts via LoRA before merging them, without continual pre-training. Synthesized hard negatives yield the largest gains for smaller models, and STM achieves strong performance on biomedical retrieval tasks while maintaining competitive general-domain results across all four backbone families.
\end{abstract}

\section{Introduction}

Retrieval-augmented generation (RAG) \cite{lewis2020rag} has become a 
standard approach for grounding Large Language Models (LLMs), improving 
on recent knowledge and mitigating hallucinations 
\cite{xiong-etal-2024-medrag,fan2024ragsurvey,abo2025ragsurvey,
ayala-bechard-2024-reducing-hallucinations,barry-etal-2025-graphrag}. 
At the core of RAG systems are lexical and semantic retrievers 
\cite{sawarkar2024blendedrag,wang2024searchingbestpracticesrag}, 
with dense retrievers built on decoder-only LLMs now achieving 
state-of-the-art performance on embedding tasks 
\cite{wang-etal-2024-improving-text,behnamghaderllm2vec}. This 
suggests that general-purpose LLMs already provide a strong 
foundation for retrieval.

\begin{figure}[t]
    \centering
    \includegraphics[width=\linewidth,trim={0.25in 0.2in 0.2in 0.2in},clip]{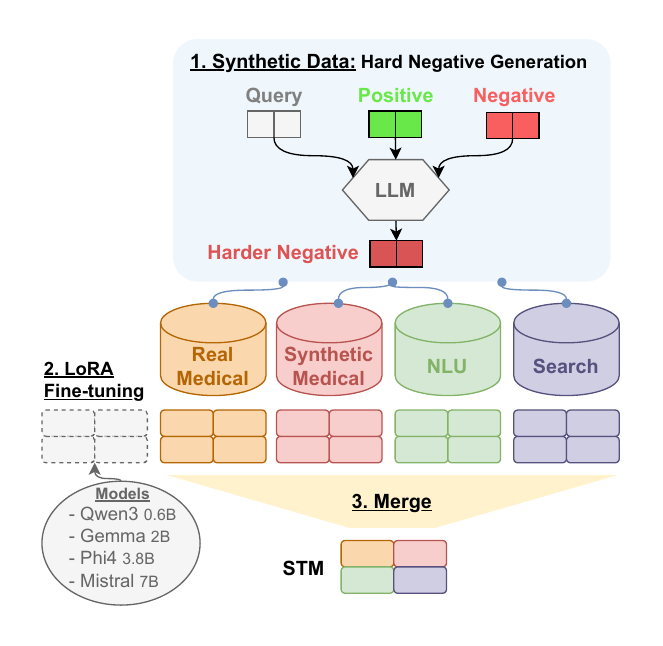}
    \caption{Diagram of our recipe to obtain the \textit{STM} retrievers: \textbf{1)}~synthetic data based on hard negative generation, \textbf{2)}~LoRA fine-tuning, and \textbf{3)}~model merging. We segment the BMRetriever dataset into four splits: \textit{Real Medical}, \textit{Synthetic Medical}, \textit{NLU}, and \textit{Search}.}
    \label{fig:main_diagram}
    \vspace{-0.2in}
\end{figure}

Yet adapting these models to domain-specific retrieval, such as in 
the biomedical field, remains an open challenge. Prior work 
\cite{xu2024bmretriever} shows that contrastive pre-training on 
large corpora followed by instruction tuning yields strong medical 
retrievers, and hard negative mining further boosts performance 
\cite{moreira2024nvretriever,lee2025nvembed,shao2025reasonir}. 
However, the computational cost of continual pre-training is 
substantial, and it remains unclear which subsets of the data 
mixture drive most of the gains, or whether top-tier LLMs can be 
leveraged to synthesize effective hard negatives in place of 
mined ones.

In parallel, model merging \cite{goddard2024mergekit} has emerged 
as a powerful paradigm for composing robust models 
\cite{wortsman2022modelsoup,aakanksha2024mixdatamergemodels}, enabling modular 
development \cite{corbeil-etal-2025-modular} and efficient data 
ablation \cite{na2024scalablemerging}. Basic techniques such as 
ModelSoup \cite{wortsman2022modelsoup} are now part of the training 
recipes of several recent retrievers, including Granite 
\cite{awasthy2025granite}, Llama-Embed-Nemotron-8B 
\cite{babakhin2025nvidia}, EmbeddingGemma \cite{vera2025embeddinggemma}, and Qwen3 Embedding \cite{zhang2025qwen3}. 
To our knowledge, only 
\citet{xiao2024lmcocktail} has systematically studied ModelSoup 
merging (i.e., linear merging) for both encoder and decoder retrievers. Beyond this, 
little is known about whether merging gains hold consistently 
across model scales and data mixtures, which data subsets matter 
most, and how to best optimize merging coefficients for retrieval.

\begin{figure}[t]
\centering
\includegraphics[width=0.48\textwidth]{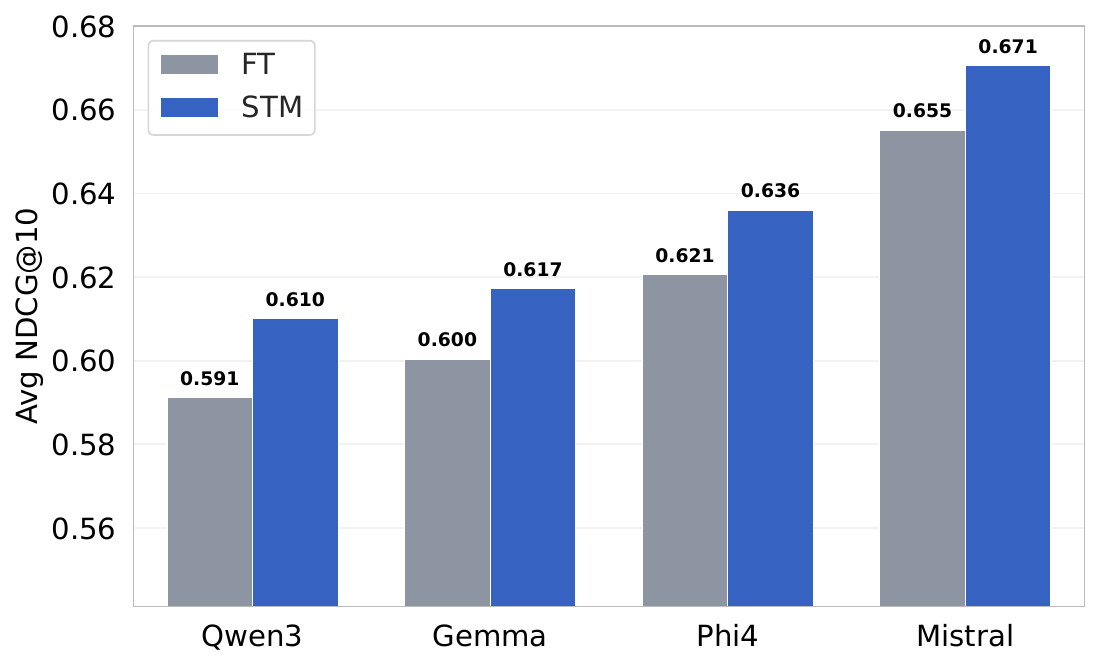}
\caption{Performance comparison of STM Merged Models versus models fine-tuned on the combined datasets of all merged experts, across 4 base models, using the average NDCG@10 metric across 12 MTEB tasks.}
\label{fig:merging-vs-finetuning-line}
\end{figure}

We ask whether merging domain-specialized experts can replace 
large-scale mixed-domain training entirely. Across four 
decoder-only LLM families and twelve retrieval tasks, we find 
consistently that it can. To maximize expert quality before 
merging, we introduce \textbf{Synthesize-Train-Merge (STM)}, 
a modular framework that synthesizes hard negatives with an LLM,
fine-tunes domain-specialized experts via LoRA, 
and merges them in parameter space; all without continual 
pre-training (Figure~\ref{fig:main_diagram}). Our contributions are as follows:

\begin{itemize}

    \item We demonstrate that parameter-space \textbf{merging 
    of dense retrieval experts consistently 
    performs above large-scale mixed-domain fine-tuning} across four decoder-only LLM families (0.6B--7B), four merging methods, and twelve retrieval tasks from MTEB (Figure~\ref{fig:merging-vs-finetuning-line}).

    \item We introduce \textbf{STM (Synthesize-Train-Merge)}, 
    a modular framework for building domain-specialized dense 
    retrievers without continual pre-training.

    \item We show the practical effectiveness of \textbf{LLM-synthesized hard negatives} for retrieval expert training across a range of backbone families, with especially strong benefits for smaller models.

    \item We release \textbf{code}
    \footnote{ 
    \url{https://github.com/TIO-IKIM/STM}}, an improved \textbf{retrieval dataset}, and \textbf{STM model checkpoints} \footnote{\url{https://huggingface.co/collections/ikim-uk-essen/stm}.} to 
    facilitate reproducibility and future research.    

\end{itemize}

\section{Related Works}

\begin{table}[t]
    \renewcommand{\arraystretch}{1.8}
    \setlength{\tabcolsep}{3.5pt}
    \centering
    \small
    \begin{tabular}{lccc}
        \toprule
         & \textbf{LLM2Vec}
         & \textbf{BMRetriever}
         & \makecell{\textbf{STM} \scriptsize{(Ours)}} \\
        \midrule
        \textbf{Pooling} 
            & Average 
            & EOS 
            & EOS \\
        \textbf{Training Setup} 
            & LoRA
            & LoRA
            & LoRA \\
        \textbf{Training Recipe} 
            & \makecell{MNTP +\\SimSCE} 
            & PT + FT 
            & \makecell{FT +\\Merging} \\
        \textbf{Negatives} 
            & SimSCE 
            & \makecell{Sampled\\Top-k} 
            & Synthetic \\
        \textbf{Dataset size} 
            & 1.5M 
            & 11.4M 
            & 1.4M \\
        \textbf{Domain} 
            & General 
            & Biomedical 
            & \makecell{General \&\\Biomedical} \\
        \bottomrule
    \end{tabular}
    \caption{Comparison of attributes between previous methods (LLM2Vec, BMRetriever) and ours. LLM2Vec is a multi-task contrastive embedding model, and BMRetriever a biomedical dense retriever pretrained with pseudo-queries (PT) and finetuned on a mix of real and synthetic labeled data with mined hard negatives (FT).}
    \label{tab:comp_models}
\end{table}

\subsection{Retrievers from Decoder-Only Models}

E5 \cite{wang-etal-2024-improving-text} first demonstrated state-of-the-art performance on the MTEB benchmark \cite{muennighoff2022mteb} from pretraining and fine-tuning Mistral 7B \cite{jiang_mistral_2023}, a decoder-only model, on real and synthetic paired retrieval data. LLM2Vec \cite{behnamghaderllm2vec} showed that using bidirectional attention masking with average pooling on LLMs, and training them with \textit{masked next token prediction} and SimCSE \cite{gao2021simcse} led to improvements on retrieval tasks. NV-Embed \cite{lee2025nvembed} introduced the \textit{latent attention} pooling layer during training with positive-aware hard negatives and non-retrieval task data to reach stronger performance. BMRetriever \cite{xu2024bmretriever} leveraged a vast contrastive pre-training, followed by instruction tuning on a mix of real and synthetic datasets, yielding strong dense retrieval models for the biomedical domain.

\subsection{Hard Negative Mining}

Hard negative mining has become a crucial design choice for training modern dense retrievers. Classical negative mining schemes \citep{xiong2021approx,zhan2021optimizing} showed that retrieving top-ranked passages during training can accelerate contrastive learning. Despite its limitation to exact word overlap, BM25 \citep{robertson2009probabilistic} is still widely used to surface hard negatives \citep{karpukhin-etal-2020-dense,zhan2021optimizing}. NV-Retriever \citep{moreira2024nvretriever} revisits this space with \emph{positive-aware} hard negative mining, which boosts the performance on the MTEB benchmark.

Beyond mined hard negatives, a small but growing line of work starts to explore \emph{generated} hard negatives using LLMs. SyNeg \citep{li2024syneg} uses an LLM with a multi-attribute self-reflection prompting strategy to synthesize hard negatives and combines them with retrieved negatives in a hybrid sampling scheme. Their ablation study shows that gains only arise from the hybrid method. In contrast, we show that only prompting a top-tier LLM to generate hard negatives can yield substantial gains, particularly for smaller models.

\subsection{Model Merging}

Linear-mode connectivity \cite{frankle2020lmc1,mirzadeh2020linearmodeconnectivitymultitask} established that independently trained solutions can be connected by low-loss paths in parameter space, motivating model combination methods based on interpolation. Building on this insight, Model Soup \cite{wortsman2022modelsoup} showed that averaging multiple training checkpoints can outperform selecting a single one. These ideas naturally extend from combining checkpoints of a single model to merging distinct expert models: \textit{task arithmetic} \cite{ilharco2023taskarithmetic} formulates merging as adding and subtracting task-specific deltas from a base model, while methods such as Ties-merging \cite{yadav2023ties} and DARE \cite{yu2024dare} explicitly tackle parameter interference when merging multiple experts. Recent work further shows that such parameter-level merging can be competitive with data-mixing strategies \cite{aakanksha2024mixdatamergemodels,na2024scalablemerging}.

Authors \cite{labrak2024biomistral,corbeil-etal-2025-modular} employed merging to build robust medical LLMs. EmbeddingGemma \cite{vera2025embeddinggemma} and Qwen3 Embedding \cite{zhang2025qwen3} exploit merging in their recipe without studying its impact.




\section{Methods}

\subsection{Hard Negative Generation}

Training dense retrievers with contrastive objectives requires negatives that are both topically related and semantically distinct from the positives. Existing hard negative mining strategies frequently struggle to balance informativeness and correctness, often introducing either trivial negatives or false negatives that are actually relevant \cite{moreira2024nvretriever}.

To alleviate this issue, we employ GPT-5 \cite{openai2025gpt5} to generate synthetic hard negatives. Given a query $q$, a positive passage $p^+$, and an existing mined negative $p^-$, we prompt the LLM to generate a new negative passage $\tilde{p}^-$ that remains lexically and topically aligned with $q$ while being semantically irrelevant or contradictory. The prompt provides the full context $(q, p^+, p^-)$ and explicitly instructs the model to preserve surface-level similarity while altering semantic intent. The exact prompt template is provided in Appendix~\ref{app:hard_negative_prompts}.




\subsection{Instruction Fine-Tuning}

We fine-tune decoder-only backbone models using a contrastive learning objective to obtain dense retrievers. We adopt the InfoNCE loss~\cite{infoNCE}, which encourages each query embedding to be closer to its corresponding positive passage than to all other passages in the batch and to any provided hard negative passage.

Formally, given a batch of $N$ triplets $\{(q_i, p_i^+,p_i^-)\}_{i=1}^N$, the loss term for a given query $q_i$ is defined as:

\[
    \mathcal{L}_i = -\log \frac{e^{\mathrm{sim}(q_i, p_i^+)}}{e^{\mathrm{sim}(q_i, p_i^-)} + \sum_{j=1}^{N} e^{\mathrm{sim}(q_i, p_j^+)}},
\]
where $\mathrm{sim}(\cdot, \cdot)$ denotes cosine similarity between embeddings, and the denominator includes two terms: the first one uses a hard negative $p_i^-$, and the second one employs other passages $p_j^+$ for $j \neq i$ as in-batch negatives. We average over the loss terms in the batch to obtain the total loss.


\subsubsection{Expert Model Fine-Tuning}

To enable modular specialization, we fine-tune multiple expert retrievers on different, coherent subsets of the BMRetriever fine-tuning dataset \cite{xu2024bmretriever}: \textit{medical synthetic}, \textit{medical real}, \textit{natural language understanding (NLU)}, and \textit{search}.

The \textit{medical real} subset includes sentence-level biomedical inference and similarity datasets such as MedNLI~\cite{shivade2017mednli} and Medical Question Pairs~\cite{mccreery2020mqp}, as well as passage-level biomedical QA benchmarks including MEDIQA~\cite{benabacha2019mediqa}, medical StackExchange QA~\cite{flax2021stackexchangeqp}, and medical dialogue data~\cite{li2023chatdoctor}. The \textit{medical synthetic} subset consists of LLM-generated biomedical retrieval pairs. To improve general-domain relevance modeling, the dataset further incorporates \textit{NLU} benchmarks such as Natural Questions~\cite{kwiatkowski2019natural}, FEVER~\cite{NanoFeverthorne-etal-2018-fever}, ELI5~\cite{fan2019eli5}, SNLI~\cite{bowman2015snli}, and the MS MARCO passage ranking dataset~\cite{bajaj2016msmarco} as the \textit{search} subset.

We train four experts per backbone model, each emphasizing a distinct data composition or training configuration, such as synthetic hard negatives.

\subsection{Model Merging}


Model merging aims to combine multiple expert retrievers into a single model that inherits complementary strengths without additional training. Given a set of expert models $\{M_k\}_{k=1}^K$ with parameters $\{\theta_k\}$, merging methods compute a unified model $\hat{M}$ by operating directly in parameter space.

In the simplest case, linear merging~\cite{wortsman2022modelsoup} is defined as:

\begin{equation}
    \theta_{\hat{M}} = \sum_{k=1}^{K} \alpha_k \theta_k,
\end{equation}
where $\alpha_k$ are the \textit{weight} coefficients with $0 \leq \alpha_k \leq 1$. 

The task arithmetic approach \cite{ilharco2023taskarithmetic} relies instead on a linear combination of task vectors $\tau_k=\theta_k - \theta_B$, which is the delta of parameters between the $k^{th}$ expert and the parameters of the base model $\theta_B$. The merged model becomes

\begin{equation}
\theta_{\hat{M}} = \theta_B + \sum_{k=1}^{K} \alpha_k \tau_k,
\end{equation}

Ties merging~\cite{yadav2023ties} also leverages task vectors $\tau_k$ with two strategies to mitigate the \textit{task-interference} phenomenon: keeping only high-magnitude parameter changes by introducing a second parameter named density $\delta_k \in [0, 1]$, and the sign agreement algorithm, which is a majority vote on the signs across all $\tau_k$. The DARE-Ties technique \cite{yu2024dare} replaces the magnitude-based pruning of Ties with random dropout at rate $\delta_k$ followed by a rescaling of task vectors.

\section{Experiments}

\subsection{Finetuning Expert Datasets}
For fine-tuning, we reuse the BMRetriever data mixture \cite{xu2024bmretriever} which originally contained 14 data sources. We reorganize it into the four coherent subsets in Table~\ref{tab:dataset_split_counts} balanced by size, task, and domain.

\begin{table}[!h]
    \centering
    \resizebox{\linewidth}{!}{%
        \begin{tabular}{lcc|cc}
            \toprule
                &  \multicolumn{2}{c}{\textbf{Medical}}  & \multicolumn{2}{c}{\textbf{General}} \\
            \midrule
            \textbf{Split}& Med-Synth & Med-Real  & Search & NLU \\
            \midrule
             \textbf{\#Pairs} & 431,000   & 306,000 & 438,000 & 251,000 \\
            \bottomrule
        \end{tabular}
    }
    \caption{Pair counts for our custom splits of the BMRetriever fine-tuning dataset, comprising four splits: two in the medical domain and two in the general domain.}
    \label{tab:dataset_split_counts}
    \vspace{-0.2in}
\end{table}

\subsection{Training Setup}

We experiment with four decoder-only backbone architectures of varying sizes; Qwen3~\cite{yang2025qwen3}, Gemma~\cite{team2024gemma}, Phi-4~\cite{abdin2024phi4}, and Mistral~\cite{jiang_mistral_2023}; the configurations are summarized in Table~\ref{tab:models}.

\begin{table}[h!]
    \small
    \renewcommand{\arraystretch}{1.3}
    \centering
    \begin{tabular}{lcccc}
        \toprule
         & \textbf{Qwen3} & \textbf{Gemma} & \textbf{Phi4} & \textbf{Mistral}  \\
         \midrule
        \textbf{Parameters} & 0.6B & 2B & 3.8B & 7B \\
        \textbf{Dimensions} & 1,024 & 2,048 & 3,072 & 4,096 \\
        \bottomrule
    \end{tabular}
    \caption{Backbone model used for expert training. \textit{Parameters} indicate the total model size, and \textit{Dimensions} refers to the hidden size.}
    \label{tab:models}
\end{table}

In preliminary experiments, we consider both EOS-token pooling and mean pooling strategies. However, we observe that the former yields slightly stronger performance. Therefore, we adopt EOS pooling for all reported results. Table \ref{tab:comp_models} compares our implementations with previous works.

During fine-tuning, retrieval prompts are prepended to each query, and models are trained using the InfoNCE loss. We fine-tune all models using LoRA adapters~\cite{hu2022lora} applied to all linear layers following prior work \cite{behnamghaderllm2vec,lee2025nvembed,xu2024bmretriever}. Hyperparameters, including learning rate, batch size, number of steps, and LoRA configuration, are summarized in Table~\ref{tab:ft_params} of Appendix~\ref{sec:appendix}. 

We conduct fine-tuning under several configurations. We begin with standard fine-tuning on the base dataset and the four data splits. We then fine-tune the models on the datasets augmented with synthetic hard negatives. Finally, we evaluate a configuration that combines the best experts from either standard fine-tuning or synthetic hard-negative variants.

\subsection{Model Merging}

We perform model merging using MergeKit~\cite{goddard2024mergekit}, which supports parameter-space merging for HuggingFace-compatible models. We evaluate four main merging strategies: linear interpolation~\cite{wortsman2022modelsoup}, task arithmetic~\cite{ilharco2023taskarithmetic}, Ties merging~\cite{yadav2023ties}, and DARE-Ties~\cite{yu2024dare}. 

We compare the selection of the best merged models across two different coefficient-optimization techniques: (1) a \textit{random} search approach of $N$ sampling rounds, and (2) the CMA-ES \textit{evolutionary} algorithm \cite{akiba2025evolutionary} at $N$ evaluations based on the MergeKit implementation\footnote{Adapted for the evaluation of retrieval tasks.}. For linear merging and task arithmetic, we sweep weight coefficients $\alpha \in \{0, 0.1, \dots, 0.9\}$. For Ties and DARE-Ties, we sweep the weight coefficients over the same range, and vary the density parameter over $\rho \in \{0.1, 0.2, \dots, 0.9\}$.

Without any training, all merged models are evaluated on a frozen development
split comprising four retrieval tasks: NFCorpus~\cite{NFCCorpusboteva2016},
FiQA-2018~\cite{FIQA2018thakur2021beir}, FEVER~\cite{NanoFeverthorne-etal-2018-fever},
and CLUE-IR~\cite{corbeil-etal-2025-modular}, spanning biomedical in-domain
and general out-of-domain retrieval. For NFCorpus and FiQA-2018, dev queries
are sampled from the official validation split and are disjoint from the full
test corpora used in MTEB evaluation; FEVER and CLUE-IR are held exclusively
for tuning and do not appear in the benchmark. Hyperparameters are selected
using unweighted mean NDCG@10 across the four subsets. Full details of the
frozen development benchmark construction, sampling procedure, search
configuration, and merge-selection protocol are provided in
Appendix~\ref{par:frozen-dev} and Appendix~\ref{app:merge-parameters}.

\subsection{Baselines}

We compare our models against a diverse set of retrieval baselines, including BM25~\cite{bm25s}, Contriever \cite{izacard2021contriever}, E5-v2~\cite{wang-etal-2024-improving-text}, GTR~\cite{gtrni2021largedualencodersgeneralizable}, LLM2Vec 3B~\cite{behnamghaderllm2vec}, BMRetriever 2B~\cite{xu2024bmretriever}, and Promptriever 7B~\cite{weller2024promptrieverinstructiontrainedretrieversprompted}. Baselines are selected to match our models in parameter scale or architectural family and to reflect comparable training budgets. 

\subsection{Evaluation}

Previous works have focused on either the medical domain or the general domain. In this work, we aim at maximizing medical performance while maintaining general retrieval capabilities. We evaluate retrieval performance on the English medical subset of the Medical MTEB benchmark~\cite{muennighoff2022mteb, enevoldsen2025mmtebmassivemultilingualtext}, which includes TREC-COVID~\cite{TRECCOVIDroberts2021searching}, SciFact~\cite{wadden-etal-2020-fact}, NFCorpus~\cite{NFCCorpusboteva2016}, CUREv1~\cite{Athar_Sheikh_2025}, PublicHealthQA~\cite{pubhealthqaxing_han_lu_2024}, FeedbackQA ~\cite{FEEDBACKQAli-etal-2022-using} and MedicalQA~\cite{MedicalQABenAbacha-BMC-2019}. To assess general-domain generalization, we additionally evaluate on five general-domain MTEB datasets, including FiQA~\cite{FIQA2018thakur2021beir}, ArguAna~\cite{Arguanawachsmuth2018retrieval}, SciDocs~\cite{SCIFACTspecter2020cohan}, and two NanoMTEB subsets (NanoFEVER~\cite{NanoFeverthorne-etal-2018-fever} and NanoQuora~\cite{Nanoquora-question-pairs}).

Following prior work on dense retriever evaluation~\cite{xu2024bmretriever,weller2024promptrieverinstructiontrainedretrieversprompted, muennighoff2022mteb}, we report NDCG@10 and Recall@100 on established public benchmarks. Scores are averaged across three evaluation groups: medical datasets, general-domain datasets, and the full dataset collection. For instruction-tuned retrievers, we use a shared retrieval prompt across all models at evaluation time. 

Descriptions of the twelve tasks, including corpus sizes and query types, are in Appendix~\ref{app:task-descriptions}.

\section{Results and Analysis}

\subsection{Comparison with Prior Retrievers}

\begin{table}[h]
\centering
\small
\renewcommand{\arraystretch}{1.3}
\setlength{\tabcolsep}{6pt}
\begin{tabular}{l c c c c}
\toprule
\textbf{Model} & \textbf{Size} & \makecell{Avg\\Med} & \makecell{Avg\\General} & \makecell{Avg\\All} \\
\midrule
BM25 & - & 0.532 & 0.515 & 0.525 \\
Contriever & 150M & 0.508 & 0.533 & 0.519 \\
E5 \scriptsize{Large V2} & 335M & 0.654 & 0.576 & 0.622 \\
GTR \scriptsize{T5 XL} & 1.2B & 0.581 & 0.586 & 0.583 \\
BMRetriever & 2B & 0.645 & 0.560 & 0.609 \\
LLM2Vec & 3B & 0.635 & 0.597 & 0.619 \\
Promptriever & 7B & 0.682&	0.605&	0.650 \\
\midrule
$STM_{Qwen3}$ & 0.6B & 0.636 & 0.574 & 0.610 \\
$STM_{Gemma}$ & 2B & 0.648 & 0.573 & 0.617\\
$STM_{Phi4}$ & 3.8B & 0.659 & 0.604 & 0.636 \\
$STM_{Mistral}$ & 7B & \textbf{0.693} & \textbf{0.639} & \textbf{0.671} \\
\bottomrule
\end{tabular}
\caption{Summary of retrieval performance against baselines across medical and general domains. NDCG@10 scores are averaged across 12 MTEB tasks. Bold indicates the highest value per column. Full per-task results are reported in Appendix~\ref{app:further_results}, Table~\ref{tab:model_comparison_biomed}.}
\label{tab:model_comparison_summary}
\vspace{-0.1in}
\end{table}

We evaluate the best-merged STM models against established baselines on twelve MTEB tasks (Table~\ref{tab:model_comparison_summary}), following established practice in retrieval evaluation \cite{muennighoff2022mteb,enevoldsen2025mmtebmassivemultilingualtext,xu2024bmretriever}.

$STM_{Mistral}$ records the highest mean NDCG@10 across the twelve tasks (0.671), 0.021 above the strongest baseline, Promptriever 7B (0.650), and 0.049 above E5 Large V2 (0.622).
BMRetriever shows higher medical than general scores (0.645/0.560), indicating biomedical specialization at the cost of general-domain coverage. STM models show a more balanced profile across both subsets. $STM_{Gemma}$ records a higher general score than BMRetriever (0.573 vs.\ 0.560) despite sharing the same 2B base model, without sacrificing medical performance (0.648 vs.\ 0.645). $STM_{Qwen3}$ (0.6B) records higher mean NDCG@10 than GTR T5 XL (1.2B) (0.610 vs.\ 0.583). Recall@100 results are consistent with these findings (Table~\ref{tab:model_comparison__recall_summary}).


\subsection{Model Merging}

\begin{table}[h]
\centering
\small
\renewcommand{\arraystretch}{1.12}
\setlength{\tabcolsep}{5pt}
\begin{tabular}{lcccc}
\toprule
\textbf{Method} 
& \textbf{Qwen3} 
& \textbf{Gemma} 
& \textbf{Phi4} 
& \textbf{Mistral} \\
\midrule
\multicolumn{5}{c}{Full Dataset Fine-Tuning} \\
\cmidrule(lr){1-5}
FT-all$_{HNM}$ & 0.591 & 0.600 & 0.621 & 0.655 \\
FT-all$_{SHN}$ & 0.600 & 0.605 & 0.629 & 0.648 \\
\midrule
\multicolumn{5}{c}{Modular Specialized Experts} \\
\cmidrule(lr){1-5}
Best Expert$_{HNM}$ & 0.583 & 0.591 & 0.611 & 0.652 \\
Best Expert$_{SHN}$ & 0.592 & 0.598 & 0.605 & 0.647 \\
\midrule
\multicolumn{5}{c}{Merged Experts (STM)} \\
\cmidrule(lr){1-5}
STM-DARE & 0.608 & 0.615 & \textbf{0.636} & 0.660 \\
STM-TIES & 0.608 & \textbf{0.617} & 0.630 & \textbf{0.671} \\
STM-Linear & \textbf{0.610} & 0.608 & 0.623 & 0.662 \\
STM-TA & 0.607 & \textbf{0.617} & 0.634 & 0.668 \\
\bottomrule
\end{tabular}
\caption{Comparison of full-dataset fine-tuning, modular specialized experts, and STM-based expert merging across four backbone families. NDCG@10 scores are averaged across 12 MTEB tasks. Bold indicates the best-performing result within each backbone family. Full results are reported in Appendix~\ref{app:further_results}, Table~\ref{tab:full_variants_with_experts}.}
\label{tab:avg_all_onecol}
\vspace{-0.2in}
\end{table}

Table~\ref{tab:avg_all_onecol} shows that merged models record higher mean NDCG@10 than fully fine-tuned counterparts across all backbone families and model sizes (Figure~\ref{fig:merging-vs-finetuning-line}). TIES and Task Arithmetic achieve the same average NDCG@10 across the four families (0.632), with TIES leading for Mistral (0.671), Task Arithmetic tying with TIES for Gemma (0.617), and DARE-TIES leading for Phi4 (0.636). This observation suggests mixed results with respect to merging techniques; while simple weighted arithmetic on task vectors is sufficient in certain settings, more complex interference-aware methods offer an advantage in others. Interference-aware methods (TIES, DARE-TIES) are preferable when the risk of task interference is non-negligible, consistent with the high sign-conflict rate we measure across expert task vectors (\textasciitilde 87\% across all four backbones; Table~\ref{tab:sign_conflict}), which motivates their use as the number of experts or interference grows.

In all four families, the best STM variant records higher mean NDCG@10 than both the strongest individual expert and full-dataset fine-tuning, confirming that merging captures complementary domain strengths rather than amplifying a single dominant expert. 

Figure~\ref{fig:models-grid} in Appendix~\ref{app:further_results} offers a mechanistic perspective on the observed benchmark advantage of merging over full-data fine-tuning. Full-data fine-tuning (dashed gray) produces larger per-layer weight deviations from the base model than any individual expert, suggesting it overwrites rather than preserves domain-specific structure. The best merged model (solid red) instead stays closer to its expert parents in parameter space, inheriting their complementary specializations without the destructive interference that joint training introduces. This pattern is consistent across all four backbone families.

\subsection{Synthetic Hard Negatives}
\label{sec:shn}

\subsubsection{Expert-level Ablations}

\begin{table}[h]
\centering
\small
\renewcommand{\arraystretch}{1.10}
\setlength{\tabcolsep}{5pt}
\begin{tabular}{lccccc}
\toprule
\textbf{Model} 
& \textbf{\makecell{Med-\\Synth}} 
& \textbf{\makecell{Med-\\Real}} 
& \textbf{Search} 
& \textbf{NLU} 
& \textbf{Avg} \\
\midrule
\multicolumn{6}{c}{Qwen3 0.6B} \\
\cmidrule(lr){1-6}
FT$_{HNM}$ & 0.583 & 0.568 & 0.506 & 0.546 & 0.551 \\
FT$_{SHN}$ & 0.592 & 0.583 & 0.557 & 0.563 & 0.574 \\
$\Delta_{\%}$ & +1.5\% & +2.6\% & +10.1\% & +3.1\% & +4.2\% \\
\midrule
\multicolumn{6}{c}{Gemma 2B} \\
\cmidrule(lr){1-6}
FT$_{HNM}$ & 0.581 & 0.591 & 0.503 & 0.520 & 0.549 \\
FT$_{SHN}$ & 0.598 & 0.575 & 0.508 & 0.559 & 0.560 \\
$\Delta_{\%}$ & +2.9\% & -2.7\% & +1.0\% & +7.5\% & +2.0\% \\
\midrule
\multicolumn{6}{c}{Phi4 3.8B} \\
\cmidrule(lr){1-6}
FT$_{HNM}$ & 0.611 & 0.611 & 0.497 & 0.585 & 0.576 \\
FT$_{SHN}$ & 0.605 & 0.600 & 0.591 & 0.566 & 0.590 \\
$\Delta_{\%}$ & -1.0\% & -1.8\% & +18.9\% & -3.2\% & +2.4\% \\
\midrule
\multicolumn{6}{c}{Mistral 7B} \\
\cmidrule(lr){1-6}
FT$_{HNM}$ & 0.638 & 0.652 & 0.610 & 0.603 & 0.626 \\
FT$_{SHN}$ & 0.644 & 0.647 & 0.571 & 0.620 & 0.621 \\
$\Delta_{\%}$ & +0.9\% & -0.8\% & -6.4\% & +2.8\% & -0.8\% \\
\bottomrule
\end{tabular}
\caption{NDCG@10 scores averaged over 12 MTEB tasks for expert models trained with standard hard-negative mining (HNM) and synthetic hard negatives (SHN). $\Delta_{\%}$ denotes the relative percentage improvement of SHN over HNM.}
\label{tab:experts_variants_vertical}
\vspace{-0.1in}
\end{table}

In Table~\ref{tab:experts_variants_vertical}, the synthetic hard negative approach (SHN) yields consistent improvements over standard hard-negative mining (HNM) for three of the four backbones, with average NDCG@10 gains of +4.2\%, +2.0\%, and +2.4\% for Qwen3-0.6B, Gemma-2B, and Phi4-3.8B, respectively. These gains are driven mainly by the general-domain Search split (+10.1\% for Qwen3, +18.9\% for Phi4), while the biomedical splits show mixed or negative changes. Mistral-7B is the sole exception, showing a slight average decline of −0.8\%, suggesting that larger, higher-capacity models may already learn sufficiently discriminative representations from mined hard negatives, leaving less room for synthetic augmentation to contribute.

An inverse scaling trend also emerges: smaller models derive the largest average benefit from SHN (+4.2\% for Qwen3-0.6B versus −0.8\% for Mistral-7B). We hypothesize that smaller encoders, having more limited representational capacity, are more sensitive to the quality of negative examples during contrastive training. Our results are consistent with recent findings on scaling laws for dense retrieval \citep{fang2024scaling}, which show that model size and training data quality interact predictably, though the specific interaction with synthetic hard negatives warrants further study.

We verified a stratified sample of the SHN dataset using an independent embedding model not used in STM training. GPT-5 negatives occupy the expected hard-negative regime (Table~\ref{tab:similarity_distributions}): their mean query--passage cosine similarity ($0.56$) is comparable to mined negatives ($0.54$) and well below positives ($0.73$), while random negatives fall to $0.20$. A two-judge false-negative audit with two LLMs distinct from the generator (Table~\ref{tab:hn_audit}) confirms 95--96\% of GPT-5 negatives as topically related to the query but not correct answers, with 96\% raw inter-judge agreement.

\subsubsection{Ablation of Modular Expert Selection}
\label{sec:expert-pool-mteb}

\begin{table}[t]
\centering
\small
\setlength{\tabcolsep}{5pt}
\begin{tabular}{@{}lcccc@{}}
\toprule
Pool & Qwen3 & Gemma & Phi4 & Mistral \\
\midrule
\texttt{base} & \textbf{0.607} & 0.614 & 0.625 & 0.667 \\
\texttt{shn}  & 0.605 & 0.589 & 0.616 & 0.653 \\
\texttt{best} & 0.604 & \textbf{0.617} & \textbf{0.634} & \textbf{0.668} \\
\bottomrule
\end{tabular}
\caption{Mean NDCG@10 for Task Arithmetic merging across 
expert pools. \texttt{base}: standard 
experts; \texttt{shn}: synthetic hard-negative experts; 
\texttt{best}: architecture-specific mix.}
\label{tab:expert-pool-ta}
\vspace{-0.3in}
\end{table}

Table~\ref{tab:expert-pool-ta} shows that the \texttt{best} pool, a curated mix of standard and SHN adapters, records the highest mean NDCG@10 for Gemma (0.617), Phi4 (0.634), and Mistral (0.668). This is particularly notable for Phi4 and Mistral, where SHN provided little or no improvement at the single-expert level (Table~\ref{tab:experts_variants_vertical}), yet SHN adapters contribute meaningfully when composed via merging. The reverse holds for Qwen3: despite benefiting most from SHN at the single-expert level, all pools perform within 0.003 points of each other at merging, consistent with the inverse scaling trend in Table~\ref{tab:experts_variants_vertical}, suggesting that at 0.6B, the base experts already capture sufficient complementarity for merging to exploit. Full results across all methods are in Table~\ref{tab:expert-pool-by-method}.

\subsection{Merging Coefficients}
\subsubsection{Optimization of Coefficients}
\label{sec:merge-params-compare-results}

Table~\ref{tab:merge-params-by-method} compares CMA-ES evolve and random search across all four model families. The two strategies achieve comparable peak performance, and neither dominates consistently across architectures or methods. Random search with $N{=}10$ samples is therefore a competitive and substantially cheaper alternative. For Phi4, random search records higher mean NDCG@10 than evolve on three of four methods (e.g., DARE-TIES: 0.636 vs.\ 0.624 for evolve), which may reflect the higher-dimensional weight space making CMA-ES convergence harder within 50 evaluations. For Mistral, random search records higher mean NDCG@10 for Task Arithmetic 
(0.668 vs.\ 0.653) and TIES (0.671 vs.\ 0.665). These results suggest that practitioners can obtain near-optimal merged models with a lightweight random sweep over a coarse coefficient grid, without requiring evolutionary optimization.

\subsubsection{Analysis of Coefficients}
\label{sec:merge_coef}

Merge weight coefficients for the best configuration per method and backbone are depicted in Figure~\ref{fig:expert-weights}\footnote{Density coefficients for TIES are excluded as weight coefficients directly modulate expert amplitude.} of Appendix \ref{app:merge-parameters}. Medical experts receive the highest weights across all families, with MedReal consistently dominant. As detailed in Table~\ref{tab:merge-params-by-method}, several optimal configurations assign near-zero weight to at least one expert: NLU is down-weighted to 0.01 in Gemma's best Task Arithmetic configuration (MedReal~0.84, Search~0.41), Search to 0.06 in Phi4's Task Arithmetic base configuration (MedReal~0.43), and Search to 0.01 in Mistral's Task Arithmetic hn configuration (MedReal~0.41, MedSynth~0.44). The redundant expert varies by architecture, confirming that the optimal subset emerges naturally from the search process rather than requiring manual selection.

\begin{figure}[h!]
    \centering
    \includegraphics[width=0.9\linewidth]{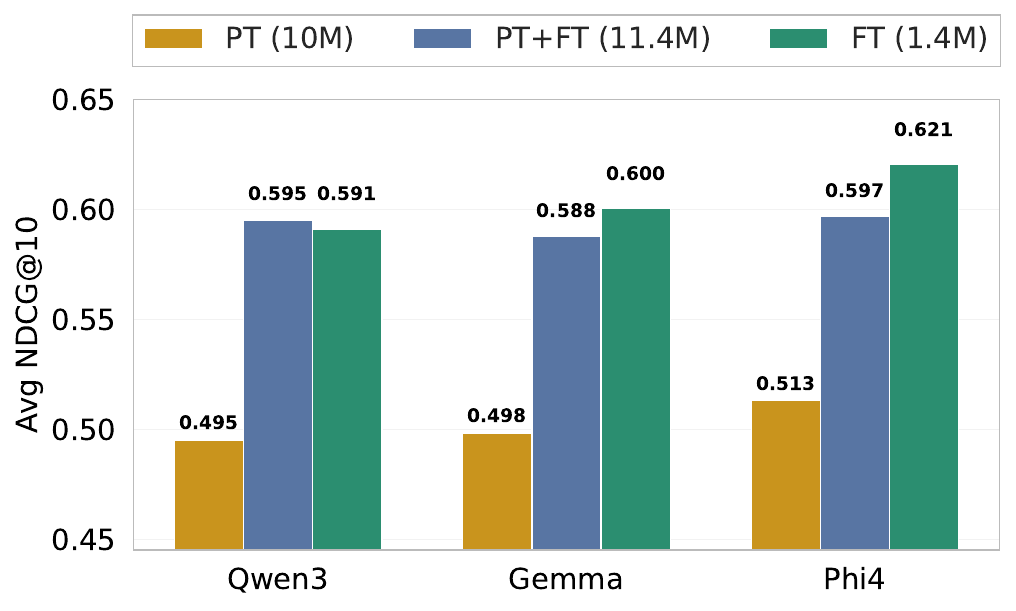}
    \caption{Performance averages of three base models pre-trained (PT) and fine-tuned (FT) on BMRetriever of 10M and 1.4M samples, respectively.}
    \label{fig:pt_ft_comp}
    \vspace{-0.2in}
\end{figure}

\subsection{What about pretraining?}

We displayed in Figure \ref{fig:pt_ft_comp} results comparing pretraining (PT) and fine-tuning (FT) setups. Despite several previous works on both general~\cite{wang-etal-2024-improving-text,lee2025nvembed} and medical domains~\cite{xu2024bmretriever} employing an expansive pretraining phase, we observe that pretraining on 10M unlabeled pairs underperforms models fine-tuned only on 1.4M pairs. We therefore drop the pretraining step and use fine-tuning only in our experiments, reducing the training compute of a single fine-tuning run by approximately 88\% relative to a PT+FT pipeline (Table~\ref{tab:compute_comparison}).

\section{Discussion}

\paragraph{Architecture vs. scale.}
A confounding factor of our setup is that each scale is represented by a single architecture, which prevents disentangling the effects of model family and size. However, our main findings are consistent across all four backbone families, indicating that overall the gains from expert merging are not architecture-specific. Rather than attributing improvements to scale, we interpret these results as evidence of robust cross-architecture behavior. While we study multiple scales, evaluating several sizes within a single family would more precisely disentangle scaling effects for one fixed architecture, which we leave to future work.

\paragraph{On the role and cost of modularity.} A natural baseline to model merging is to fine-tune a single model on the union of all data. However, our results consistently show that merging specialized experts performs above full-data fine-tuning, suggesting that joint training tends to average or overwrite domain-specific signals, while modular training preserves them and enables their effective composition. We use identical fine-tuning hyperparameters per backbone in both settings (Table~\ref{tab:ft_params}), and as shown in Table~\ref{tab:base_only_merge}, this effect is not attributable to our GPT-5 synthetic hard negatives: merging base experts without SHN still records higher mean NDCG@10 than full-data fine-tuning across all four backbones. Both settings still include the LLM-generated Med-Synth pairs from the original BMRetriever mixture, so the comparison isolates the contribution of our GPT-5 hard negatives rather than of LLM-generated data in general. This supports a view of merging not as added complexity, but as a way to better retain complementary expertise.

\begin{table}[t]
\centering
\footnotesize
\setlength{\tabcolsep}{4pt}
\begin{tabular}{@{}lccc@{}}
\toprule
Family & FT-all & \makecell{Base\\Merge} &  \makecell{Merge\\Method} \\
\midrule
Qwen3 0.6B  & 0.591 & \textbf{0.610} & Linear     \\
Gemma 2B    & 0.600 & \textbf{0.615}  & DARE-TIES  \\
Phi4 3.8B   & 0.621 & \textbf{0.636}  & DARE-TIES  \\
Mistral 7B  & 0.655 & \textbf{0.667}  &Task-Arith \\
\bottomrule
\end{tabular}
\caption{Mean NDCG@10 across 12 MTEB tasks for merged models built exclusively from experts trained on mined negatives (Base Merge), compared against full-data fine-tuning on the same data (FT-all). Full per-task results are reported in Appendix~\ref{app:further_results}, Table~\ref{tab:base_and_hn_merge_results}.}
\label{tab:bas}
\label{tab:base_only_merge}
\vspace{-0.1in}
\end{table}

In practice, the added complexity is minimal: merging is performed on CPU offline on a small development set, using simple coefficient search, and requires no additional training. The training compute for the four experts is approximately equal to that of a single full-data fine-tuning run, since the splits cover essentially the same $\sim$1.4M-pair dataset. The observed variation in wall-clock across backbones (Table~\ref{tab:wallclock_compute}) reflects implementation-level throughput rather than a systematic difference. Compared to rank-based fusion methods~\cite{cormack2009reciprocal}, the key distinction is at inference time: ensembling requires running multiple models, whereas merging produces a single model, yielding a better efficiency–performance trade-off.

\paragraph{Why multiple experts.}
Finally, the use of multiple experts reflects complementary signals (real, synthetic, NLU, search), and our results indicate that combining them improves performance over both individual experts and joint training, suggesting that these signals are not redundant but synergistic.

\section{Conclusion}

We showed that merging domain-specialized dense retrieval 
experts consistently performs above large-scale mixed-domain 
training across four decoder-only LLM families (0.6B--7B), 
four merging methods, and twelve MTEB tasks, suggesting 
that parameter-space composition captures complementary 
domain strengths that mixed-domain training averages out. 
STM synthesizes hard negatives with a top-tier LLM, 
fine-tunes domain-specialized experts via LoRA, and merges 
them in parameter space without continual pre-training. 
Synthesized hard negatives yield the largest gains for 
smaller models, and optimal merge configurations naturally 
down-weight redundant experts, confirming that merging 
exploits complementarity rather than averaging it out. 
These results suggest that modular expert composition is 
a practical and effective alternative to mixed-domain 
training for building versatile dense retrievers.

\section*{Limitations}
Our study has three main limitations. First, our synthetic hard negative generation relies on one proprietary LLM; we do not evaluate sensitivity to the choice of generator model, and performance may vary with open-weight alternatives, though we note that hard negative generation is a one-time offline step that does not affect inference cost. Second, all fine-tuning and merging runs use a single random seed per configuration; multi-seed evaluation could further quantify variability. Third, our experiments are conducted exclusively on English retrieval tasks; extending STM to multilingual or cross-lingual settings is an important direction given the growing need for domain-specialized retrieval in non-English biomedical literature.

\section*{Acknowledgments}

We thank Fran\c{c}ois Beaulieu and Paul Vozila for their contributions to earlier versions of this work. We acknowledge the use of AI-assisted tools to improve writing clarity and assist with coding. All AI-assisted outputs were reviewed, edited, and validated by the authors. This publication was supported by the German Federal Ministry of Education and Research (BMBF) Network of University Medicine 2.0: "NUM 2.0", Grant No. 01KX2121, Project: RACOON-BI.

\bibliography{custom}

\appendix

\newpage

\section{Implementation Details}
\label{sec:appendix}

\begin{table}[!h]
    \small
    \renewcommand{\arraystretch}{1.3}
    \centering
    \begin{tabular}{lc}
        \toprule
        \textbf{Hyperparameter} &  \\
        \midrule
        Maximum Tokens & 512 \\
        Optimizer & AdamW \\
        LR Scheduler & Linear Warmup \\
        \# Warmup Steps & 100 \\
        bf16 & True \\
        Learning Rate & \\
        \quad Mistral Instruct v0.2 (7B) & $1\ \times\ 10 ^{-5}$ \\
        \quad Phi4 mini instruct (3.8B) & $2\ \times\ 10 ^{-5}$ \\
        \quad Gemma (2B) & $1\ \times\ 10 ^{-5}$ \\
        \quad Qwen3 (0.6B) & $5\ \times\ 10 ^{-5}$ \\
        \# Epochs & 1 \\
        LoRA Rank & 16 \\
        LoRA $\alpha$ & 32 \\
        Total Batch Size & \\
        \quad Mistral Instruct v0.2 (7B) & 32 \\
        \quad Phi4 mini instruct (3.8B) & 64 \\
        \quad Gemma (2B) & 128 \\
        \quad Qwen3 (0.6B) & 256 \\
        \bottomrule
    \end{tabular}
    \caption{Fine-tuning Hyperparameters.}
    \label{tab:ft_params}
\end{table}

\begin{table}[h]
\centering
\small
\renewcommand{\arraystretch}{1.3}
\setlength{\tabcolsep}{6pt}
\begin{tabular}{l c c c c}
\toprule
\textbf{Model} & \textbf{Size} & \makecell{Avg\\Med} & \makecell{Avg\\General} & \makecell{Avg\\All} \\
\midrule
BM25 & - & 0.650 & 0.774 & 0.702 \\
Contriever & 150M & 0.664 & 0.798 & 0.720 \\
E5 \scriptsize{Large V2} & 335M & 0.724 & 0.830 & 0.769 \\
GTR \scriptsize{T5 XL} & 1.2B & 0.690 & 0.814 & 0.742 \\
BMRetriever & 2B & 0.691 & 0.768 & 0.723 \\
LLM2Vec & 3B & 0.730 & 0.846 & 0.778 \\
Promptriever & 7B &0.729&	0.831&	0.771\\
\midrule
$STM_{Qwen3}$ & 0.6B & 0.713 & 0.811 & 0.754 \\
$STM_{Gemma}$ & 2B & 0.709 & 0.786 & 0.741 \\
$STM_{Phi4}$ & 3.8B & 0.728 & 0.844 & 0.776 \\
$STM_{Mistral}$ & 7B & \textbf{0.741} & \textbf{0.856} & \textbf{0.790} \\
\bottomrule
\end{tabular}
\caption{Retrieval performance measured in Recall@100 across medical and general domains. \textbf{Bold} indicates the best result per column.}
\label{tab:model_comparison__recall_summary}
\vspace{-0.1in}
\end{table}

\newpage
\section{Prompt Examples}
\label{app:prompts}

\subsection{Hard Negative Generation Prompts}
\label{app:hard_negative_prompts}

The following prompt template was used for hard negative mining:

\begin{promptbox}[Hard Negative Generation Prompt]
\small 
Given this query: \texttt{"\{query\}"}

Original prompt: \texttt{"\{original prompt\}"}  

Original positive document: \texttt{"\{positive doc\}"}

Original negative document: \texttt{"\{original negative\}"}

Generate a HARD negative document that is:

1. Related to the same domain (extract domain from the query)

2. Contains similar terminology and concepts as the positive document

3. But is NOT relevant to answering the specific query

4. Should be moderately challenging to distinguish from the positive document

5. Should be a realistic document in the domain

6. Should be harder than the original negative document but not as hard as a super hard negative

IMPORTANT: Return ONLY the document text. Do not include any introductory text, explanations, summaries, or meta-commentary. Just return the raw document content.

\end{promptbox}

\begin{figure}[ht]
\centering
\begin{subfigure}[t]{0.48\linewidth}
\centering
\includegraphics[width=\linewidth]{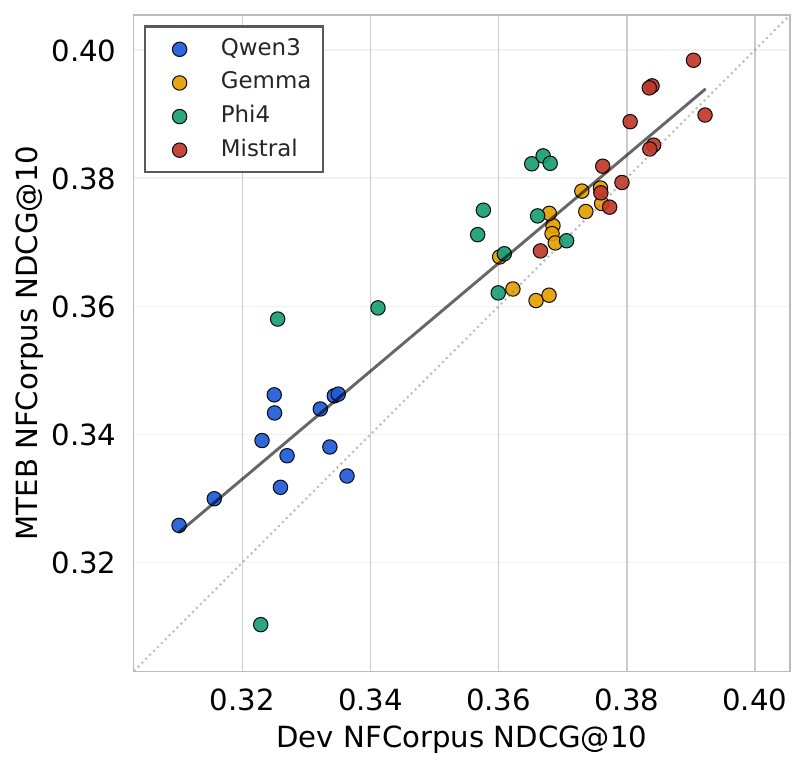}
\caption{NFCorpus (medical). Pearson $r=+0.93$, Spearman $\rho=+0.91$.}
\label{fig:appendix-dev-nfc}
\end{subfigure}
\hfill
\begin{subfigure}[t]{0.48\linewidth}
\centering
\includegraphics[width=\linewidth]{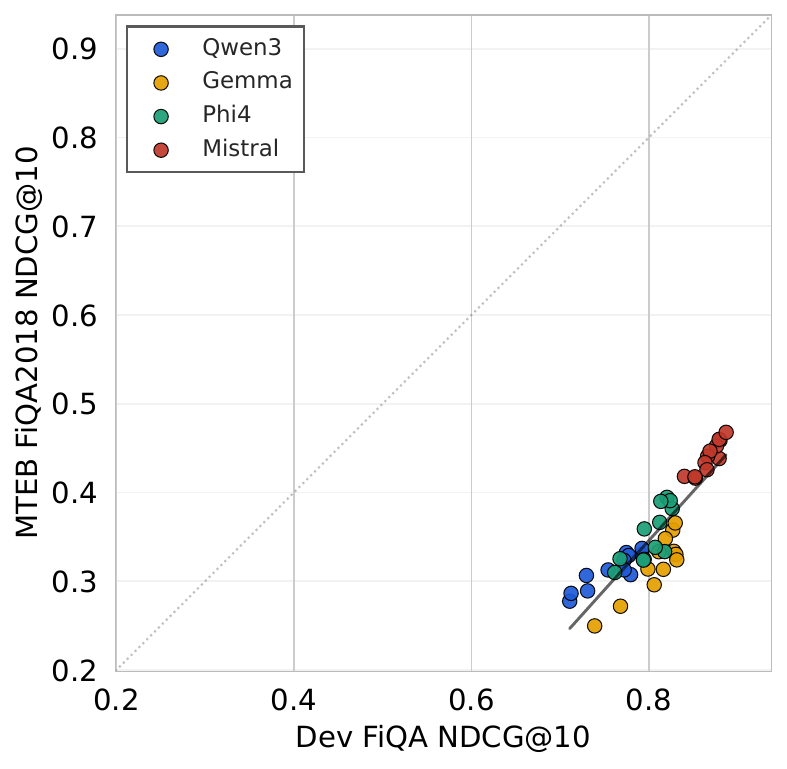}
\caption{FiQA2018 (general). Pearson $r=+0.89$, Spearman $\rho=+0.89$.}
\label{fig:appendix-dev-fiqa}
\end{subfigure}
\caption{Same-task dev sampling validation (125-query dev NDCG@10 vs.\ full test benchmark). Each point is one random-search merged model (Qwen3, Gemma, Phi4, Mistral; twelve merge-method variants per architecture). Dotted: identity; solid: linear fit.}
\label{fig:appendix-dev-sampling}
\end{figure}

\begin{figure*}[t]
  \centering
  \includegraphics[width=\textwidth]{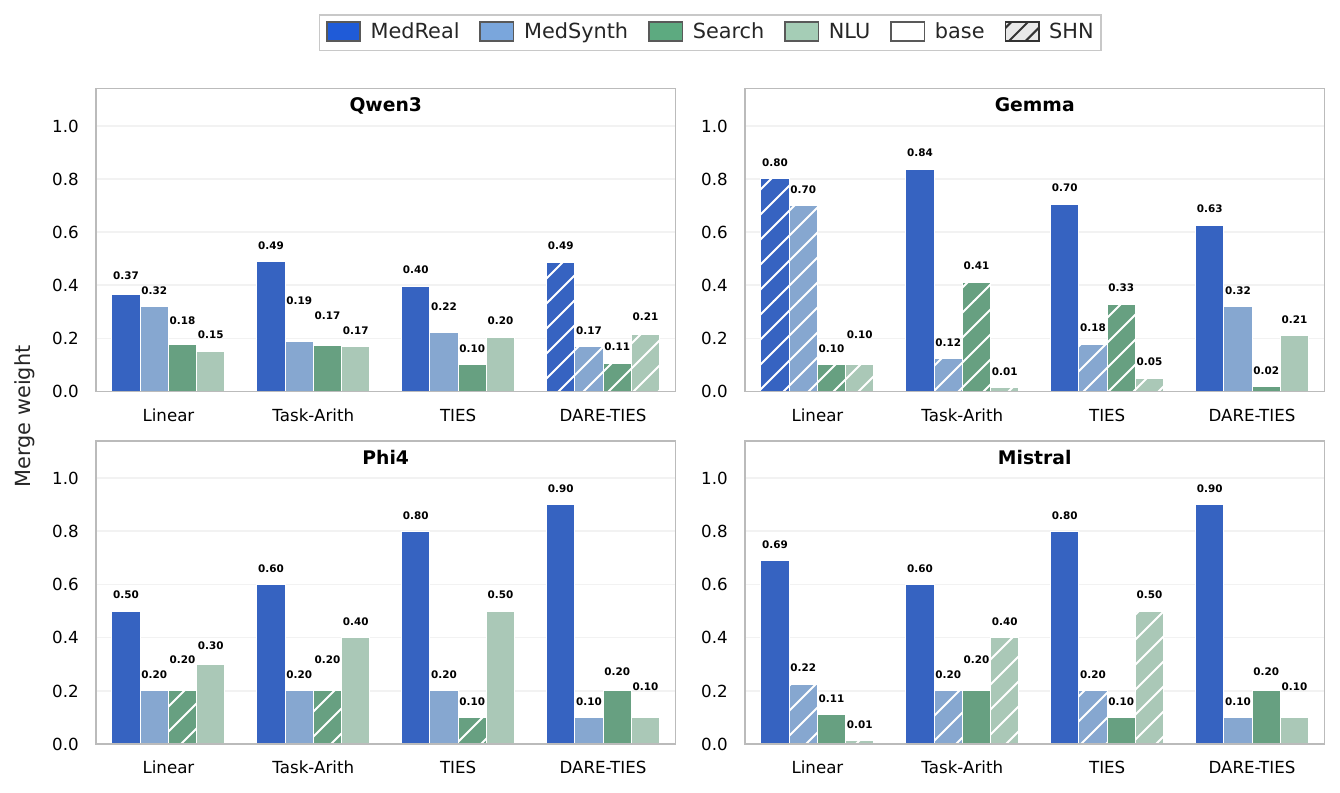}
  \caption{%
    \textbf{Expert merge weights} for the best merged model per merging method
    on each backbone (Qwen3, Gemma, Phi4, Mistral).
    Each panel shows parent coefficients assigned to four domain experts:
    MedReal, MedSynth,
    Search, and NLU.
    For every backbone and method, we select the merged checkpoint with the
    highest \texttt{mean NDCG@10} over the benchmark suite.
    Solid bars use the standard expert; hatched bars (\texttt{//}) use the
    synthetic hard-negative (SHN) variant.
    Bar labels report merge weights.
  }
  \label{fig:expert-weights}
\end{figure*}

\begin{figure*}[t]
  \centering
  \includegraphics[width=\textwidth]{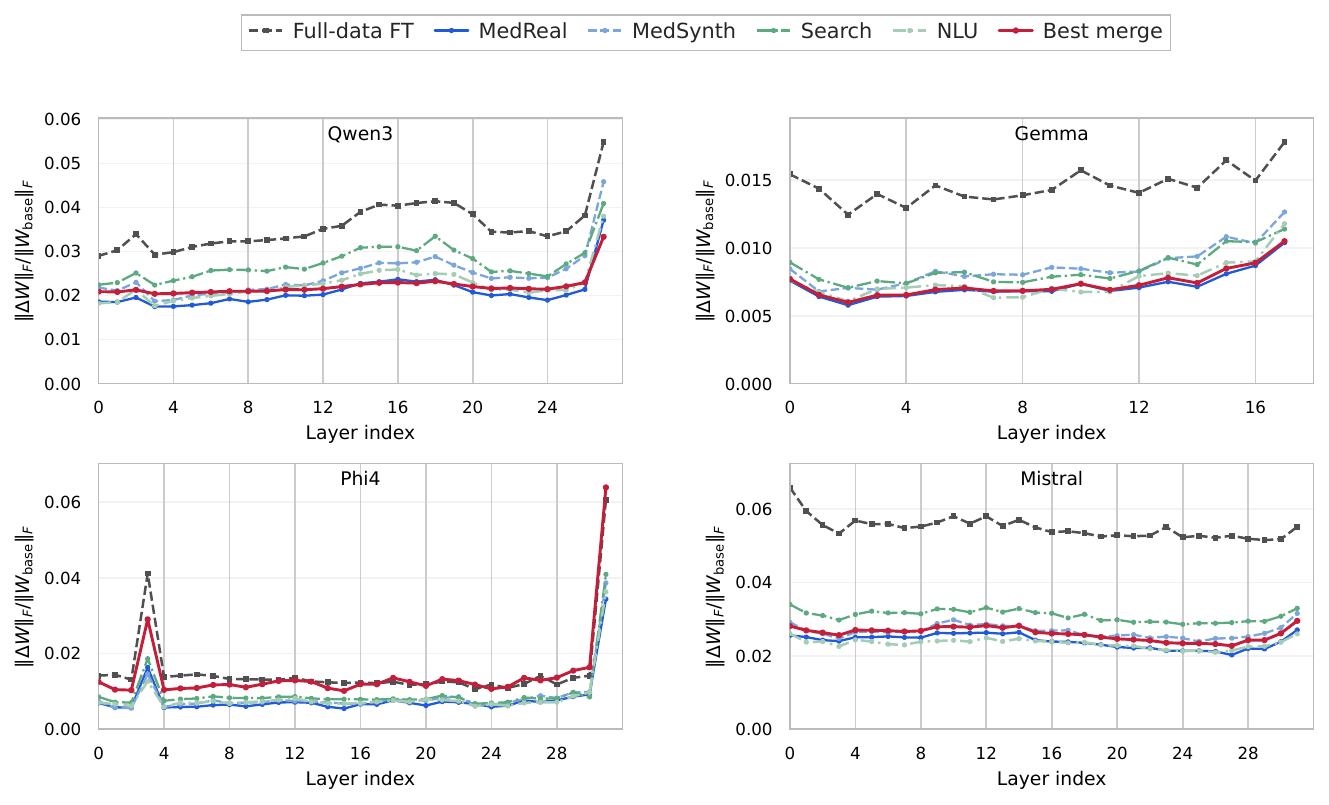}
  \caption{Per-layer relative Frobenius weight change
    $\|\Delta W\|_F / \|W_{\mathrm{base}}\|_F$ for all four 
    backbone families (Qwen3, Gemma, Phi4, Mistral), where 
    $W_{\mathrm{base}}$ is the pretrained base checkpoint.
    Dashed gray: full-data fine-tuning. Solid red: best 
    merged model (highest mean NDCG@10 across 12 MTEB tasks per family). 
    Colored curves: individual domain experts used in that 
    merged checkpoint.}
  \label{fig:models-grid}
\end{figure*}

\section{Detailed Results}
\label{app:further_results}

\begin{table*}[t]
\centering
\small
\renewcommand{\arraystretch}{1.1}
\setlength{\tabcolsep}{5pt}

\begin{tabular}{l|ccc|ccc|ccc|ccc}
\toprule
\textbf{Method}
& \multicolumn{3}{c|}{\textbf{Qwen3}}
& \multicolumn{3}{c|}{\textbf{Gemma}}
& \multicolumn{3}{c|}{\textbf{Phi4}}
& \multicolumn{3}{c}{\textbf{Mistral}} \\
\cmidrule(lr){2-4} \cmidrule(lr){5-7} \cmidrule(lr){8-10} \cmidrule(lr){11-13}

& \makecell{Avg\\Med} & \makecell{Avg\\Gen} & \makecell{Avg\\All}
& \makecell{Avg\\Med} & \makecell{Avg\\Gen} & \makecell{Avg\\All}
& \makecell{Avg\\Med} & \makecell{Avg\\Gen} & \makecell{Avg\\All}
& \makecell{Avg\\Med} & \makecell{Avg\\Gen} & \makecell{Avg\\All} \\
\midrule

\multicolumn{13}{c}{Full Dataset Fine-Tuning} \\
\cmidrule(lr){1-13}

FT-all$_{HNM}$
& 0.622 & 0.548 & 0.591
& 0.638 & 0.548 & 0.600
& 0.655 & 0.573 & 0.621
& 0.692 & 0.603 & 0.655 \\

FT-all$_{SHN}$
& 0.632 & 0.555 & 0.600
& 0.637 & 0.561 & 0.605
& 0.661 & 0.585 & 0.629
& 0.676 & 0.609 & 0.648 \\

\midrule
\multicolumn{13}{c}{Modular Specialized Experts} \\
\cmidrule(lr){1-13}

Med-Real
& 0.609 & 0.510 & 0.568
& 0.625 & 0.542 & 0.591
& 0.643 & 0.567 & 0.611
& 0.678 & 0.616 & 0.652 \\

Med-Real$_{SHN}$
& 0.613 & 0.541 & 0.583
& 0.609 & 0.527 & 0.575
& 0.626 & 0.564 & 0.600
& 0.676 & 0.606 & 0.647 \\

Med-Synth
& 0.614 & 0.540 & 0.583
& 0.606 & 0.546 & 0.581
& 0.642 & 0.567 & 0.611
& 0.671 & 0.592 & 0.638 \\

Med-Synth$_{SHN}$
& 0.614 & 0.561 & 0.592
& 0.624 & 0.562 & 0.598
& 0.640 & 0.555 & 0.605
& 0.670 & 0.608 & 0.644 \\

Search
& 0.527 & 0.477 & 0.506
& 0.523 & 0.475 & 0.503
& 0.515 & 0.472 & 0.497
& 0.638 & 0.571 & 0.610 \\

Search$_{SHN}$
& 0.577 & 0.530 & 0.558
& 0.510 & 0.507 & 0.508
& 0.617 & 0.555 & 0.591
& 0.598 & 0.533 & 0.571 \\

NLU
& 0.573 & 0.509 & 0.546
& 0.546 & 0.484 & 0.520
& 0.628 & 0.525 & 0.585
& 0.623 & 0.575 & 0.603 \\

NLU$_{SHN}$
& 0.578 & 0.543 & 0.563
& 0.564 & 0.552 & 0.559
& 0.585 & 0.539 & 0.566
& 0.637 & 0.596 & 0.620 \\

\midrule

\multicolumn{13}{c}{Merged Experts (STM)} \\
\cmidrule(lr){1-13}

$STM_{\text{DARE}}$
&0.631&\textbf{0.577}&0.608
&\textbf{0.649}&0.567&0.615
&\textbf{0.659}&\textbf{0.604}&\textbf{0.636}
&0.684&0.626&0.660
\\

$STM_{\text{TIES}}$
&0.634&0.571&0.608
&\textbf{0.649}&0.572&\textbf{0.617}
&\textbf{0.659}&0.590&0.630
&\textbf{0.693}&\textbf{0.639}&\textbf{0.671}\\

$STM_{\text{Linear}}$
&\textbf{0.636}&0.574&\textbf{0.610}
&0.631&\textbf{0.576}&0.608
&0.648&0.589&0.623
&0.687&0.627&0.662\\

$STM_{\text{TA}}$
&0.632&0.572&0.607
&0.648&0.573&\textbf{0.617}
&0.661&0.595&0.634
&\textbf{0.693}&0.633&0.668\\
\bottomrule

\end{tabular}

\caption{
Average NDCG@10 across 12 MTEB tasks. Specialized experts are trained on partitioned subsets of the training data, while FT-all denotes fine-tuning on the full dataset.
}

\label{tab:full_variants_with_experts}
\vspace{-0.1in}
\end{table*}

\begin{table*}[t]
\centering
\small
\renewcommand{\arraystretch}{1.1}
\setlength{\tabcolsep}{1.7pt}
\begin{tabular}{lc|ccccccc|ccccc|cc|c}
\toprule
& & \multicolumn{7}{c|}{\scriptsize \underline{\textbf{Medical}}} & \multicolumn{5}{c|}{\scriptsize \underline{\textbf{General}}} & & & \\
\textbf{\scriptsize Model} & \textbf{\scriptsize Size} & \makecell{\scriptsize Feedback\\\scriptsize QA} & \makecell{\scriptsize Medical\\\scriptsize QA} & \makecell{\scriptsize CUREv1} & \makecell{\scriptsize Public\\\scriptsize HealthQA} & \makecell{\scriptsize NF\\\scriptsize Corpus} & \makecell{\scriptsize TREC\\\scriptsize COVID} & \makecell{\scriptsize Sci\\\scriptsize Fact} & \makecell{\scriptsize SCI\\\scriptsize DOCS} & \makecell{\scriptsize Nano\\\scriptsize FEVER} & \makecell{\scriptsize Argu\\\scriptsize Ana} & \makecell{\scriptsize Nano\\\scriptsize Quora} & \makecell{\scriptsize FiQA\\\scriptsize 2018} & \makecell{\scriptsize\textbf{Avg}\\\scriptsize\textbf{Med.}} & \makecell{\scriptsize\textbf{Avg}\\\scriptsize\textbf{Gen.}} & \makecell{\scriptsize\textbf{Avg}\\\scriptsize\textbf{All}} \\
\midrule
BM25 & - & 0.563 & 0.458 & 0.355 & 0.718 & 0.321 & 0.623 & 0.686 & 0.158 & 0.809 & 0.492 & 0.863 & 0.251 & 0.532 & 0.515 & 0.525 \\
Contriever & 150M & 0.505 & 0.592 & 0.351 & 0.694 & 0.313 & 0.448 & 0.655 & 0.171 & 0.794 & 0.484 & 0.944 & 0.274 & 0.508 & 0.533 & 0.519 \\
E5 \scriptsize{Large V2} & 335M & 0.704 & 0.699 & 0.562 & 0.856 & 0.372 & 0.666 & 0.722 & 0.205 & 0.889 & 0.464 & 0.912 & 0.411 & 0.654 & 0.576 & 0.622 \\
GTR \scriptsize{T5 XL} & 1.2B & 0.577 & 0.692 & 0.507 & 0.713 & 0.333 & 0.601 & 0.642 & 0.157 & 0.846 & 0.528 & 0.957 & 0.442& 0.581 & 0.586 & 0.583 \\
BMRetriever & 2B & 0.587 & 0.727 & 0.471 & 0.812 & 0.347 & \textbf{0.839 }& 0.729 & 0.186 & \textbf{0.940 }& 0.356 & 0.960 & 0.357 & 0.645 & 0.560 & 0.609 \\
LLM2Vec & 3B & 0.708 & 0.731 & 0.490 & 0.814 & \textbf{0.385 }& 0.572 & 0.746 & 0.190 & 0.864 & 0.553 & 0.953 & 0.423 & 0.635 & 0.597 & 0.619 \\
Promptriever & 7B & 0.714	&0.734&	0.574&	0.855&	0.368&	0.769	&0.759&	0.185&	0.896&	0.526&	0.967&	0.451&	0.682&	0.605&	0.650
\\
\midrule
$STM_{Qwen3}$& 0.6B & 0.662	& 0.702	&0.473&	0.821&	0.346&	0.762	&0.688&	0.191	&0.824	&0.567	&0.959	&0.327	&0.636	&0.574	&0.610 \\
$STM_{Gemma}$ & 2B &0.629	&0.728&	0.439	&0.878	&0.379	&0.759	&0.728	&0.205&	0.840	&0.534	&0.954	&0.335&	0.648	&0.573&	0.617 \\
$STM_{Phi4}$ & 3.8B & 0.695&	0.747&	0.475&	0.858&	0.382&	0.721&	0.732&	\textbf{0.225}	&0.880	&0.558&	\textbf{0.964}	&0.394	&0.659	&0.604	&0.636 \\
$STM_{Mistral}$ & 7B & \textbf{0.737}&	\textbf{0.782}&	\textbf{0.571}&	\textbf{0.895}&	\textbf{0.385}&	0.732&	\textbf{0.751}&	0.222&	\textbf{0.924}&	\textbf{0.629}&	0.951&	\textbf{0.468}&	\textbf{0.693}&	\textbf{0.639}&	\textbf{0.671}\\
\bottomrule
\end{tabular}
\caption{Comparison of retrieval performance on English tasks from MTEB. Results are reported primarily on the medical subset, with additional evaluation on five general-domain subsets. Performance is measured using NDCG@10.}
\label{tab:model_comparison_biomed}
\end{table*}

\subsection{Frozen development benchmark}
\label{par:frozen-dev}

To avoid overfitting merge hyperparameters to the twelve-task MTEB benchmark,
we construct a frozen dev split: 500 queries (125 per subset, seed~42) from
four retrieval tasks spanning biomedical in-domain retrieval (NFCorpus,
CLUE-IR) and general out-of-domain settings (FiQA2018, FEVER).
Hyperparameters are selected by unweighted mean NDCG@10 across the
four subsets (Table~\ref{tab:frozen-dev-breakdown}). For NFCorpus and
FiQA2018, dev queries are sampled from the official validation split and are
disjoint from the full test corpora used in MTEB evaluation; CLUE-IR and
FEVER are held exclusively for tuning and do not appear in the benchmark.

\paragraph{Dev-set validity.}
Dev scores correlate strongly with full MTEB rankings on both overlapping
tasks across $n{=}48$ random-search checkpoints (four architectures
$\times$ twelve merge-method variants per architecture; Evolve models excluded,
as they optimize directly on dev):
Pearson $r{=}{+}0.93$ [$+0.89$, $+0.96$] on NFCorpus and
$r{=}{+}0.89$ [$+0.82$, $+0.94$] on FiQA2018.
See Figure~\ref{fig:appendix-dev-sampling}.
Table~\ref{tab:random-search-sweep} reports the gap between the best sampled configuration and the arithmetic mean across the $N{=}10$ random-search draws for each backbone model. Linear merging exhibits a small gap (at most 0.015 across all backbones), indicating that most sampled configurations lie close to the best. Interference-aware methods show wider gaps, up to 0.125 for Phi4 DARE-TIES, reflecting a broader range of sampled configurations.

\subsection{Evaluation Benchmark Description}
\label{app:task-descriptions}

Table~\ref{tab:task_descriptions} describes the twelve retrieval tasks used for evaluation, drawn from the MTEB benchmark~\cite{muennighoff2022mteb}. All tasks are English passage retrieval tasks evaluated with NDCG@10, spanning biomedical and general domains with diverse query types and corpus sizes.

\begin{table}[h]
\centering
\small
\renewcommand{\arraystretch}{1.3}
\setlength{\tabcolsep}{4pt}
\begin{tabular}{llrr}
\toprule
\textbf{Dataset} & \textbf{Domain} & \textbf{\#Q} 
& \textbf{\#Docs} \\
\midrule
TREC-COVID  & Biomedical & 50   & 171K  \\
NFCorpus    & Biomedical & 323  & 3.6K  \\
SciFact     & Biomedical & 300  & 5K    \\
CUREv1      & Clinical   & 2K   & 245K    \\
PubHealthQA & Biomedical & 172  & 172   \\
MedicalQA   & Biomedical & 2048 & 2048  \\
FeedbackQA  & Biomedical & 1.9K & 2.4K  \\
FiQA-2018   & Finance    & 648  & 57K   \\
ArguAna     & General    & 1.4K & 8.7K  \\
SciDocs     & Science    & 1K   & 25K   \\
NanoFEVER   & General    & 50  & 5K    \\
NanoQuora   & General    & 50  & 5K  \\
\bottomrule
\end{tabular}
\caption{Twelve MTEB evaluation tasks. All tasks use NDCG@10. \#Q\,=\,number of queries, \#Docs\,=\,corpus size (approximate).}
\label{tab:task_descriptions}
\end{table}

\subsection{Merge parameters}
\label{app:merge-parameters}

\begin{table}[t]
\centering
\footnotesize
\setlength{\tabcolsep}{3pt}
\begin{tabular}{@{}l l r r l@{}}
\toprule
Subset & Domain & Queries & Docs & Split \\
\midrule
NFCorpus & Biomedical & 125 & 2{,}588 & dev \\
CLUE-IR  & Biomedical & 125 & 140     & dev \\
FiQA     & General    & 125 & 380     & dev \\
FEVER    & General    & 125 & 202     & dev \\
\midrule
\textbf{Total} & & \textbf{500} & \textbf{3{,}310} & \\
\bottomrule
\end{tabular}
\caption{Frozen dev split used for hyperparameter selection (125 queries per
subset, seed~42; corpora contain all relevant documents plus 20\% hard
negatives).}
\label{tab:frozen-dev-breakdown}
\end{table}

We tune merge weights on a shared frozen development benchmark (Table~\ref{tab:frozen-dev-breakdown}) using two complementary search strategies:
\emph{random search} (uniform sampling + best-of-$N$) and
\emph{evolve} (CMA-ES).
Both use the same twelve variant buckets per architecture:
four merge methods
(Linear, Task Arithmetic, TIES, DARE-TIES)
$\times$
three expert pools:
\texttt{base} (four standard fine-tuned experts),
\texttt{hn} (four hard-negative experts),
and \texttt{best} (a per-architecture mix of standard and HN adapters derived from our strongest evolve runs).

\paragraph{Random search.}
For each bucket we draw $N{=}10$ merge configurations with seed~42:
weights (and TIES/DARE-TIES densities) are sampled from
$\{0.1, 0.2, \ldots, 1.0\}$, this yields $12 \times 10 = 120$ candidates per architecture; we keep the checkpoint with highest mean dev NDCG@10 per bucket (12 final models per architecture).

\paragraph{Evolve (CMA-ES).}
We run \texttt{mergekit-evolve} with up to 50 function evaluations per bucket.
Each evaluation merges four experts, scores the result on the frozen dev split,
and returns the mean of four dev NDCG@10 scores to CMA-ES.
For Qwen3, \texttt{best} uses the same four HN adapters as \texttt{hn}, but
we still run separate buckets and sample independent weights in random search.

Table~\ref{tab:merge-params-by-method} lists, for each model family, the single best merged checkpoint under \emph{evolve} (CMA-ES on the dev set) and under \emph{random search} (best of ten sampled configurations per expert bucket), each ranked by the mean NDCG@10 in 4 dev tasks.

\begin{table}[t]
\centering
\small
\setlength{\tabcolsep}{4pt}
\begin{tabular}{@{}l l r r r r@{}}
\toprule
Method & Pool & \textbf{Qwen3} & \textbf{Gemma} 
& \textbf{Phi4} & \textbf{Mistral} \\
\midrule
\multirow{3}{*}{DARE-TIES}
 & \texttt{base} & 0.603 & \textbf{0.615} 
 & \textbf{0.636} & \textbf{0.660} \\
 & \texttt{shn}  & \textbf{0.608} & 0.604 
 & 0.612 & 0.655 \\
 & \texttt{best} & 0.606 & \textbf{0.615} 
 & 0.616 & 0.659 \\
\midrule
\multirow{3}{*}{Linear}
 & \texttt{base} & \textbf{0.610} & 0.605 
 & 0.621 & 0.657 \\
 & \texttt{shn}  & 0.607 & \textbf{0.608} 
 & 0.610 & 0.643 \\
 & \texttt{best} & 0.606 & 0.603 
 & \textbf{0.623} & \textbf{0.662} \\
\midrule
\multirow{3}{*}{Task-Arith}
 & \texttt{base} & \textbf{0.607} & 0.614 
 & 0.625 & 0.667 \\
 & \texttt{shn}  & 0.605 & 0.589 
 & 0.616 & 0.653 \\
 & \texttt{best} & 0.604 & \textbf{0.617} 
 & \textbf{0.634} & \textbf{0.668} \\
\midrule
\multirow{3}{*}{TIES}
 & \texttt{base} & \textbf{0.608} & 0.615 
 & 0.609 & 0.658 \\
 & \texttt{shn}  & 0.603 & 0.611 
 & 0.623 & 0.656 \\
 & \texttt{best} & 0.606 & \textbf{0.617} 
 & \textbf{0.630} & \textbf{0.671} \\
\bottomrule
\end{tabular}
\caption{Mean NDCG@10 by expert pool and merge method 
(max over CMA-ES evolve and random search). Bold: best 
pool per column within each method block. \texttt{base}: 
standard experts; \texttt{shn}: synthetic hard-negative 
experts; \texttt{best}: architecture-specific mix. 
For Qwen3$^\dagger$, \texttt{shn} and \texttt{best} 
use the same four SHN adapters but are searched 
independently.}
\label{tab:expert-pool-by-method}
\end{table}

\begin{table*}[t]
\centering
\footnotesize
\setlength{\tabcolsep}{4pt}
\begin{tabular}{l l c c c c c c c c c c c c}
\toprule
 & & \multicolumn{6}{c}{Evolve (CMA-ES)} 
   & \multicolumn{6}{c}{Random Search} \\
\cmidrule(lr){3-8} \cmidrule(lr){9-14}
Family & Method 
& Var. 
& \makecell{Med-\\Real} 
& \makecell{Med-\\Synth} 
& Search 
& NLU 
& \makecell{NDCG\\@10}
& Var. 
& \makecell{Med-\\Real} 
& \makecell{Med-\\Synth} 
& Search 
& NLU 
& \makecell{NDCG\\@10} \\
\midrule
\multirow{4}{*}{Qwen3$^\dagger$} 
 & Linear     & base & 0.37 & 0.32 & 0.18 & 0.15 & 0.610
              & hn   & 1.00 & 0.40 & 0.50 & 0.50 & 0.607 \\
 & Task-Arith & base & 0.49 & 0.19 & 0.17 & 0.17 & 0.607
              & best & 0.60 & 0.20 & 0.20 & 0.40 & 0.604 \\
 & TIES       & base & 0.40 & 0.22 & 0.10 & 0.20 & 0.608
              & best & 0.80 & 0.20 & 0.10 & 0.50 & 0.596 \\
 & DARE-TIES  & hn   & 0.49 & 0.17 & 0.11 & 0.21 & 0.608
              & base & 0.90 & 0.10 & 0.20 & 0.10 & 0.599 \\
\midrule
\multirow{4}{*}{Gemma} 
 & Linear     & base & 0.56 & 0.09 & 0.20 & 0.14 & 0.601
              & hn   & 0.80 & 0.70 & 0.10 & 0.10 & 0.608 \\
 & Task-Arith & best & 0.84 & 0.12 & 0.41 & 0.01 & 0.617
              & best & 0.70 & 0.10 & 0.30 & 0.50 & 0.616 \\
 & TIES       & best & 0.70 & 0.18 & 0.33 & 0.05 & 0.617
              & hn   & 0.90 & 0.20 & 0.30 & 0.20 & 0.611 \\
 & DARE-TIES  & base & 0.63 & 0.32 & 0.02 & 0.21 & 0.615
              & best & 0.60 & 0.50 & 0.20 & 0.20 & 0.615 \\
\midrule
\multirow{4}{*}{Phi4} 
 & Linear     & best & 0.22 & 0.31 & 0.17 & 0.31 & 0.623
              & best & 0.50 & 0.20 & 0.20 & 0.30 & 0.623 \\
 & Task-Arith & base & 0.43 & 0.25 & 0.06 & 0.24 & 0.625
              & best & 0.60 & 0.20 & 0.20 & 0.40 & 0.634 \\
 & TIES       & hn   & 0.52 & 0.27 & 0.20 & 0.05 & 0.623
              & best & 0.80 & 0.20 & 0.10 & 0.50 & 0.630 \\
 & DARE-TIES  & base & 0.37 & 0.10 & 0.24 & 0.22 & 0.624
              & base & 0.90 & 0.10 & 0.20 & 0.10 & 0.636 \\
\midrule
\multirow{4}{*}{Mistral} 
 & Linear     & best & 0.69 & 0.22 & 0.11 & 0.01 & 0.662
              & best & 1.00 & 0.30 & 0.40 & 0.10 & 0.655 \\
 & Task-Arith & hn   & 0.41 & 0.44 & 0.01 & 0.19 & 0.653
              & best & 0.60 & 0.20 & 0.20 & 0.40 & 0.668 \\
 & TIES       & best & 0.77 & 0.19 & 0.13 & 0.06 & 0.665
              & best & 0.80 & 0.20 & 0.10 & 0.50 & 0.671 \\
 & DARE-TIES  & best & 0.53 & 0.19 & 0.16 & 0.10 & 0.659
              & base & 0.90 & 0.10 & 0.20 & 0.10 & 0.660 \\
\bottomrule
\end{tabular}
\caption{Best Evolve (CMA-ES) and Random Search merged checkpoint per 
architecture and merge method. NDCG@10 is the mean across 12 MTEB 
tasks. \textbf{Var.}: expert pool, \texttt{base} (standard adapters), 
\texttt{hn} (hard-negative adapters), \texttt{best} (family-specific mix).  
$^\dagger$For Qwen3, \texttt{hn} and \texttt{best} use the same four SHN 
adapters but are searched independently.}
\label{tab:merge-params-by-method}
\end{table*}

\begin{table*}[h]
\renewcommand{\arraystretch}{1.2}
\setlength{\tabcolsep}{4pt}
\small
\centering
\begin{tabular}{lccccc}
\hline
\textbf{Model} & \textbf{Params (B)} & \textbf{Global BS} & \# \textbf{Pairs (M)} & \textbf{FLOPs (approx.)} & \textbf{$\Delta$ vs PT+FT} \\
\hline
Qwen3 (PT+FT)        & 0.6 & 256 & 11.4  & $1.26\times10^{20}$ & ---  \\
Qwen3 (FT 1.4M)     & 0.6 & 256 & 1.4   & $1.55\times10^{19}$ & $-88\%$ \\
\hline
Gemma (PT+FT)       & 2.0 & 128 & 11.4  & $4.20\times10^{20}$ & ---  \\
Gemma (FT 1.4M)     & 2.0 & 128 & 1.4   & $5.16\times10^{19}$ & $-88\%$ \\
\hline
Phi4 (PT+FT)        & 3.8 & 64  & 11.4  & $7.98\times10^{20}$ & ---  \\
Phi4 (FT 1.4M)      & 3.8 & 64  & 1.4   & $9.81\times10^{19}$ & $-88\%$ \\
\hline
Mistral (PT+FT)        & 7.0 & 32  & 11.4  & $1.47\times10^{21}$ & ---  \\
Mistral (FT 1.4M)      & 7.0 & 32  & 1.4   & $1.81\times10^{20}$ & $-88\%$ \\
\hline
\end{tabular}
\caption{Training compute comparison across models and setups. FLOPs are estimated as $3 \times 6ND$ \cite{kaplan2020scaling}, where $N$ is the number of model parameters, $D$ is the total number of tokens processed (number of pairs $\times$ tokens per pair), and the factor of $3$ accounts for the three forward passes required by the contrastive loss. $\Delta$ denotes the relative reduction in compute of FT compared to PT+FT.}
\label{tab:compute_comparison}
\end{table*}

\begin{table*}[t]
\centering
\footnotesize
\setlength{\tabcolsep}{4pt}

\begin{tabular*}{\textwidth}{@{\extracolsep{\fill}} l l cc cc cc cc}
\toprule

& &
\multicolumn{2}{c}{Qwen3} &
\multicolumn{2}{c}{Gemma} &
\multicolumn{2}{c}{Phi4} &
\multicolumn{2}{c}{Mistral} \\

\cmidrule(lr){3-4}
\cmidrule(lr){5-6}
\cmidrule(lr){7-8}
\cmidrule(lr){9-10}

Variant &
\makecell{Method} &
N@10 & R@100 &
N@10 & R@100 &
N@10 & R@100 &
N@10 & R@100 \\

\midrule

Expert MedReal
& --
& 0.568 & 0.742
& 0.591 & 0.731
& 0.611 & 0.766
& 0.652 & 0.784 \\

Expert MedSynth
& --
& 0.583 & 0.738
& 0.581 & 0.720
& 0.611 & 0.758
& 0.638 & 0.777 \\

Expert Search
& --
& 0.506 & 0.698
& 0.503 & 0.665
& 0.497 & 0.699
& 0.610 & 0.762 \\

Expert NLU
& --
& 0.546 & 0.719
& 0.520 & 0.686
& 0.585 & 0.749
& 0.603 & 0.763 \\

\midrule

Expert MedReal (HN)
& --
& 0.575 & 0.744
& 0.583 & 0.727
& 0.600 & 0.763
& 0.647 & 0.788 \\

Expert MedSynth (HN)
& --
& 0.592 & 0.742
& 0.598 & 0.731
& 0.605 & 0.763
& 0.644 & 0.780 \\

Expert Search (HN)
& --
& 0.558 & 0.728
& 0.508 & 0.668
& 0.591 & 0.753
& 0.571 & 0.729 \\

Expert NLU (HN)
& --
& 0.563 & 0.730
& 0.559 & 0.710
& 0.566 & 0.750
& 0.620 & 0.759 \\

\midrule

Merged (base experts)
& Linear
& 0.610 & 0.754
& 0.605 & 0.741
& 0.621 & 0.773
& 0.657 & 0.785 \\

Merged (base experts)
& Task-Arith
& 0.607 & 0.755
& 0.614 & 0.738
& 0.625 & 0.764
& 0.667 & 0.789 \\

Merged (base experts)
& TIES
& 0.608 & 0.755
& 0.615 & 0.743
& 0.609 & 0.765
& 0.658 & 0.786 \\

Merged (base experts)
& DARE-TIES
& 0.603 & 0.753
& 0.615 & 0.741
& 0.636 & 0.776
& 0.660 & 0.787 \\

\midrule

Merged (HN experts)
& Linear
& 0.607 & 0.752
& 0.608 & 0.739
& 0.610 & 0.768
& 0.643 & 0.785 \\

Merged (HN experts)
& Task-Arith
& 0.605 & 0.750
& 0.589 & 0.730
& 0.616 & 0.767
& 0.653 & 0.788 \\

Merged (HN experts)
& TIES
& 0.603 & 0.749
& 0.611 & 0.735
& 0.623 & 0.773
& 0.657 & 0.786 \\

Merged (HN experts)
& DARE-TIES
& 0.608 & 0.753
& 0.604 & 0.731
& 0.612 & 0.769
& 0.655 & 0.787 \\

\bottomrule
\end{tabular*}

\caption{Performance comparison across backbone models. HN denotes experts trained with GPT-5-generated hard negatives. N@10 and R@100 denote Avg NDCG@10 and Avg Recall@100, respectively.}
\label{tab:base_and_hn_merge_results}

\end{table*}



\begin{table}[t]
\centering
\footnotesize
\setlength{\tabcolsep}{4pt}
\begin{tabular}{@{}lcccc@{}}
\toprule
Variant & Qwen3 & Gemma & Phi4 & Mistral \\
\midrule
Sum of 4 experts (base) & 16.5 & 22.2 & 37.3 & 75.6 \\
Sum of 4 experts (HN)   & 15.1 & 19.9 & 31.9 & 54.9 \\
\midrule
Full-data FT (base)     & 18.7 & 24.4 & 83.6 & 68.8 \\
Full-data FT (HN)       & 33.4 & 23.3 & 36.4 & 40.5 \\
\bottomrule
\end{tabular}
\caption{Wall-clock training time (hours) on a single NVIDIA H100. 
}
\label{tab:wallclock_compute}
\end{table}

\begin{table}[t]
\centering
\footnotesize
\setlength{\tabcolsep}{4pt}
\begin{tabular}{@{}lrr@{}}
\toprule
Backbone & Params compared & Conflict rate \\
\midrule
Qwen3 0.6B  & 596{,}049{,}920   & 87.06\% \\
Gemma 2B    & 1{,}981{,}808{,}640 & 86.78\% \\
Phi4 3.8B   & 3{,}221{,}225{,}472 & 86.59\% \\
Mistral 7B  & 6{,}979{,}321{,}856 & 87.43\% \\
\bottomrule
\end{tabular}
\caption{Sign-conflict rate across expert task vectors $\tau_k = \theta_k - \theta_B$, computed as the fraction of parameter positions where at least one expert disagrees in sign. Conflict is substantial and consistent across all four families, motivating the use of interference-aware merging methods (TIES, DARE-TIES) that explicitly resolve sign disagreements via majority voting.}
\label{tab:sign_conflict}
\end{table}

\begin{table}[t]
\centering
\footnotesize
\setlength{\tabcolsep}{4pt}
\begin{tabular}{@{}lrrr@{}}
\toprule
Judge & NO (\%) & Uncertain (\%) & YES (\%) \\
\midrule
DeepSeek-V4-Flash & 96.21 & 0.40 & 3.39 \\
Qwen3.6-35B-A3B   & 95.01 & 0.20 & 4.79 \\
\bottomrule
\end{tabular}
\caption{Independent LLM audit of GPT-5 synthetic hard negatives on a stratified random sample of ($n=500$) triples. Two judges not used in STM training, DeepSeek-V4-Flash~\cite{deepseekai2026deepseekv4highlyefficientmilliontoken} and Qwen3.6-35B~\cite{qwen36_35b_a3b}, independently label each candidate as YES (false negative; a correct answer to the query), NO (valid hard negative), or Uncertain. Raw inter-judge agreement is 96.0\%, with Cohen’s $\kappa=0.48$.}

\label{tab:hn_audit}
\end{table}

\begin{table}[t]
\centering
\footnotesize
\setlength{\tabcolsep}{4pt}
\begin{tabular}{@{}lrrrr@{}}
\toprule
Candidate type   & Mean & Median & Q25 & Q75 \\
\midrule
sim(positive)         & 0.729 & 0.741 & 0.674 & 0.798 \\
sim(GPT-5 negative)   & 0.561 & 0.573 & 0.479 & 0.651 \\
sim(mined negative)   & 0.539 & 0.548 & 0.475 & 0.613 \\
sim(random negative)  & 0.202 & 0.197 & 0.123 & 0.270 \\
\bottomrule
\end{tabular}
\caption{Cosine similarity to the query across candidate types on a stratified sample of ($n{=}10{,}000$) triples, measured with Qwen3-Embedding-0.6B~\cite{zhang2025qwen3}, an embedding model not used in STM training. GPT-5 hard negatives lie slightly above mined negatives, well below positives, and far above random negatives; the expected regime for a genuine hard negative, topically related to the query but not a correct answer.}
\label{tab:similarity_distributions}
\end{table}

\begin{table}[t]
\centering
\small
\setlength{\tabcolsep}{6pt}
\begin{tabular}{@{}lcccc@{}}
\toprule
Backbone & Linear & Task-Arith & TIES & DARE-TIES \\
\midrule
Qwen3   & 0.005 & 0.064 & 0.083 & 0.087 \\
Gemma   & 0.006 & 0.036 & 0.038 & 0.029 \\
Phi4    & 0.008 & 0.095 & 0.107 & 0.125 \\
Mistral & 0.015 & 0.048 & 0.059 & 0.062 \\
\bottomrule
\end{tabular}
\caption{Sweep range of dev-set NDCG@10 across the $N{=}10$ random-search configurations per (backbone, method) bucket, aggregated across the three expert pools ($n{=}30$ per cell). Values report the gap between the best sampled configuration (Max) and the arithmetic mean of the ten. Linear merging is largely insensitive to the sampled weights ($\leq 0.015$ across backbones); interference-aware methods (TIES, DARE-TIES) show wider sweep ranges (up to 0.125), reflecting a stronger dependence on the specific weight configuration.}
\label{tab:random-search-sweep}
\end{table}

\end{document}